# AI-based Wearable Vision Assistance System for the Visually Impaired: Integrating Real-Time Object Recognition and Contextual Understanding Using Large Vision-Language Models

Mirza Samad Ahmed Baig[a,d], Syeda Anshrah Gillani[a,e], Shahid Munir Shah[a], Mahmoud Aljawarneh[b], Abdul Akbar Khan[d,c], Muhammad Hamzah Siddiqui[a]

[a]Department of Computing, Faculty of Engineering, Science, and Technology, Hamdard University, Karachi, Pakistan,
[b]Applied Science Private University Amman Jordan
[c]Argaam Investment, Riyadh, Kingdom of Saudi Arabia,
[d]Danat Fz LLC (owned by Argaam), Karachi, Pakistan,
[e]Doaz, South Korea,

**Abstract**

Visual impairment affects the ability of people to live a life like normal people. Such people face challenges in performing activities of daily living, such as reading, writing, traveling and participating in social gatherings. Many traditional approaches are available to help visually impaired people; however, these are limited in obtaining contextually rich environmental information necessary for independent living. In order to overcome this limitation, this paper introduces a novel wearable vision assistance system that has a hat-mounted camera connected to a Raspberry Pi 4 Model B (8GB RAM) with artificial intelligence (AI) technology to deliver real-time feedback to a user through a sound beep mechanism. The key features of this system include a user-friendly procedure for the recognition of new people or objects through a one-click process that allows users to add data on new individuals and objects for later detection, enhancing the accuracy of the recognition over time. The system provides detailed descriptions of objects in the user's environment using a large vision language model (LVLM). In addition, it incorporates a distance sensor that activates a beeping sound using a buzzer as soon as the user is about to collide with an object, helping to ensure safety while navigating their environment. A comprehensive evaluation is carried out to evaluate the proposed AI-based solution against traditional support techniques. Comparative analysis shows that the proposed solution with its innovative combination of hardware and AI (including LVLMs with IoT), is a significant advancement in assistive technology that aims to solve the major issues faced by the community of visually impaired people.

# 1. Introduction

Globally, more than 338 million people experience visual impairment, approximately 295 million experiencing moderate to severe visual impairment, and 43 million are blind [1, 2]. For visually impaired individuals, everyday life poses significant challenges to their mobility, independence, and social engagement [3, 4, 5]. Existing traditional assistive technology, such as white canes and guide dogs, are inadequate at communicating complicated contextual information about environment and only offer limited tactile and aural feedback [6, 7, 8]. Therefore, current solutions lack the customization, flexibility, and contextual awareness needed to support users in a variety of dynamic contexts [9, 10, 11, 12]. Hence, advanced assistive technologies are in increasing demand that can allow visually impaired people to live independently and have a thorough awareness of their surroundings [13, 14].

Luckily, AI algorithms (machine learning (ML) and computer vision (CV)), can be used to construct sophisticated assistive solutions [15]. Devices that can be equipped with cameras and AI algorithms can interpret visual data, identify objects, faces, text, and even emotions, and then communicate this information through auditory or haptic feedback [16, 17, 18]. In this way, a more detailed surrounding information can be obtained, and a detailed guided response can be achieved. Although there exist numerous AI based solutions such as intelligent guide robots and AI based wearable devices (refer to 2 Section for more detail on AI based assistive technologies), however still they lack in providing a comprehensive surroundings information to users to live independently.

This study presents a wearable vision assistance system that combines a hat-mounted camera connected to a Raspberry Pi 4 Model B with AI based large vision-language models (LVLM). use of LVLM enables the system to perform real-time object recognition along with a rich contextual understanding of the environment. Overall, this system is designed to give visually impaired people a chance to live a life like normal people with the help of a convenient auditory response. The key features of the proposed solution include:

- **Personalized Recognition Database**: A user-friendly function that allows the quick addition of new people and items to a personalized recognition database, improving accuracy and customization over time.

- **Contextual Understanding**: Employing LVLM for generating detailed explanations and relevant information about recognized items enables a deeper understanding of the environment.

- **Affordable Hardware**: Easy accessibility of the system by using affordable and easily available hardware components, such as the Raspberry Pi 4 and a standard camera attached to a hat.

For validating the presented system an extensive user testing was carried out with participants who have visual impairments to assess the system's efficiency. The findings indicate a notable improvement in user environmental awareness, social engagement, and autonomy.

The rest of the paper is organized as follows. Section 2 presents the latest literature related to the presented study. Section 3 provides the detail of the system development. It includes proposed system design, opted methodology, hardware and software description, object recognition and integration with LLMs. Section 4 provides the evaluation of the designed system through users satisfaction. Section 5 provides the details of the results achieved by the system during test environment. Section 6 provides the detailed discussion on the achieved results, along with system limitations and future improvements. Finally, Section 7 provides the conclusion of the study.

## 2. Literature Review

### 2.1. Conventional Assistive Technologies

Different traditional assistive technologies have been presented in literature to help visually impaired people to live a normal life. Wite canes are one of the most widely used mobility aids, providing tactile feedback to detect obstacles [19, 20, 21, 22]. However, due to a number of its shortcomings, including their inability to identify overhanging objects, head-level impediments, and trunks. Additionally, their insistence on making direct physical contact with the obstacles and their failure to provide guidance on how to approach them caused researchers to focus on various alternative technologies and approaches [23]. Assistant dogs have been utilized as an alternative to white canes to overcome their limitations [24, 25]. Although assistant dogs offer helpful companionship and support, but gaining their support requires a lot of training and attention [26, 27, 28]. This leads to replace assistant dogs with robotic dogs as suggested in different studies [29, 30, 31, 32]. Although, robotic dogs present a great deal in improving the quality of life of visually impaired people, however, developing robust and effective robotic mobility aids with static and dynamic settings for use in both indoor and outdoor environments has remained a difficult task to accomplish [33].

### 2.2. Sensors bases Assistive Technologies

As technology advanced, sensors based devices such as smart canes, ultrasonic obstacle detection and GPS navigation introduced to identify obstacles and provide feedback through sound [34]. Although assisting with navigation, these devices are not able to provide detailed contextual information such as aiding in shopping, which significantly limits a person's ability to interact with its environment.

### 2.3. Wearable Assistive Technologies

Important progress has been made in the development of wearable devices such as eSight and OrCam. These devices offer visual enhancement and optical character recognition (OCR) capabilities [35, 36]. Chest-mounted cameras were also used with computer vision system installed in them [37]. These devices were linked with smartphones for processing and connectivity purposes [38]. Despite these great works, the challenge remains the same as contextual understanding is a problem [39], still, the blind or visually impaired people cannot even do a little bit of shopping without the support of the others.

### 2.4. AI based Assistive Technologies

As the world evolve, assistive technologies for visually impaired people have been transformed by AI. AI technologies allow devices to analyze complicated visual data more accurately. Now, it is easy to identify objects, faces, text, and scenes using computer vision algorithms [40, 41]. Several mobile applications such as Seeing AI and Be My Eyes also use AI to recognize text and objects efficiently [42]. Despite having several advantages of using AI technologies, challenges remain in developing affordable, personalized, and specially contextually aware system to which a person can live almost a normal life.

### 2.5. Summary

The adoption of AI models with the wearable devices as well as robotic dogs is associated with a few critical trade-offs that ought to be considered. The major issue that needs to be addressed pertains to designing performance capabilities without compromising on the limited resources, as discussed by [43]. Many of these devices possess low processing power, memory and battery power, which greatly hampers their capacity to efficiently process and implement large sophis-ticated models [44]. Another important factor is an important tradeoff between precision of the results and their speed in real time tasks. As shown in [1], attaining real time response times often comes at the cost of sacrificing model precision because users want their wearable technology to provide immediate feedback. Moreover, it should be possible for users to adjust hardware as per personal needs, but current wearables lack this functionality as well – the concept of personalization is almost entirely missing in current systems, as noted in [45]. Thus, our work tries to solve these issues by implementing better models and providing more possibilities of customization.

*2.6. Limitations of the Existing Systems and Proposed Solutions*

The review of the existing literature of the assistive technologies (as presented above) suggests the following two major limitations of the state-of-the-art presented systems.

1. Insufficient contextual understanding.
2. Lack of training on the faces of the user's relatives or known persons.

To overcome these limitations, the following solutions have been implemented in the proposed system.

- Enhanced understanding of context by leveraging LVLMs, as illustrated in Figure 6 and Figure 7.

- A system for facial recognition and storage with the advanced training capabilities on the faces of the user's family members and acquaintances. This system enable users to add faces using a voice command-activated button.

Further detail about the proposed system's design is presented in the following section.

## 3. System Development

*3.1. Proposed System Design*

Figure 1 illustrates the core concept of our proposed system. It is evident from Figure 1 that the proposed system is designed to give blind and visually impaired individuals a comprehensive contextual grasp of their environment. People with vision impairments can use this system to see the labeled objects in their environment, such as a chair, basketball, computer, laptop, etc., as well can see their thorough description and context. For example, if a blind person is standing in a market, the system can guide him which item is located where along with the complete description of the nearby relevant items so that the person can easily purchase the item knowing about its complete detail as shown in Figure 6 and Figure 7. The proposed system also able to detect the faces of user's relatives and known personalities along with capability of adding new faces in the system as shown in Figures 8, 9, and 10.
  2.

*3.2. Methodology Opted*

The methodology opted in this research involves three operational systems as shown in Figure 2. One system serves as the core (if no button is pressed), continuously running, while the other two are activated upon button press. There are two buttons, a green button and a blue button. Pressing the green button prompts the system to ask the user to identify the individual in front of them through voice, storing this information for future reference by the core system. This process is outlined in Figures 8, 9, and 10. Pressing the blue button will make the system to capture an image, which is then processed by a LVLM to generate detailed information about the image and convey it to the user by voice assistance, as shown in Figures6 and 7.

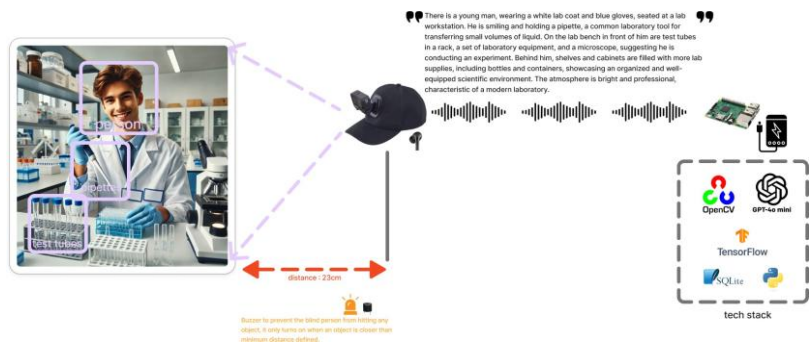

Figure 1: Overview of the proposed system

*3.3. Hardware Components*

The proposed wearable vision assistance system consists of the following hardware components (also shown in Figure 3).

- **Raspberry Pi 4 Model B (8GB RAM)**: Serves as the processing unit, chosen for its balance of performance and energy efficiency. It is capable of running AI models with great speed.

- **Camera module**: An 8MP Camera Module for Raspberry Pi, built-in to the brim of a hat, that is lightweight, and high resolution to capture photos from the subject point of view.

- **Camera support hat**: A hat that is specially designed to enable the installation of a camera.

- **Power supply**: A rechargeable battery with a capacity of 20,000 mAh is stored using a carrying case.

- **Audio output**: Stereo-acoustic headphones let sound travel directly through skull, so the user can hear clearly without covering ears. User can also hear outside of this device for staying alert about the surroundings. [46].

- **System distance calculation**: An ultrasonic sensor helps to easily determine how far away an object is by using sound waves, making it simple to determine distances.

- **Buzzer**: A buzzer is triggered to warn the user and help avoid collisions when an object is less than 20 cm.

- **User Interface Button**:
    - **Add Item Buttons**: A physical press button (green button) located on the hat enables users to input new people or objects into the system with their labels.
    - **Button for contextual details**: Another physical press button (the dark blue button) prompts the system to deliver a detailed description of the currently focused object using LLM.

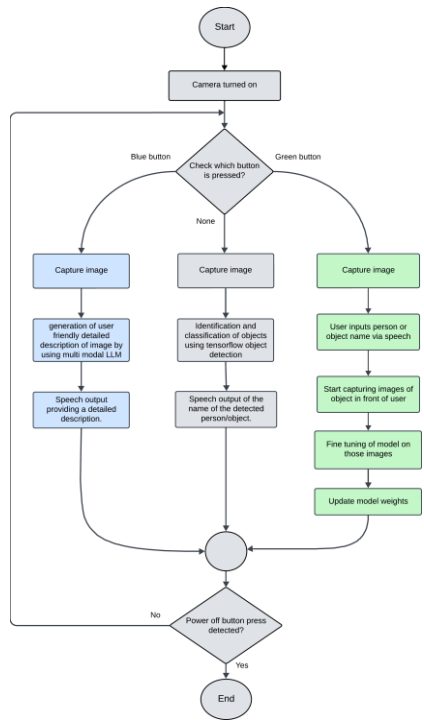

Figure 2: Methodology of the proposed system design

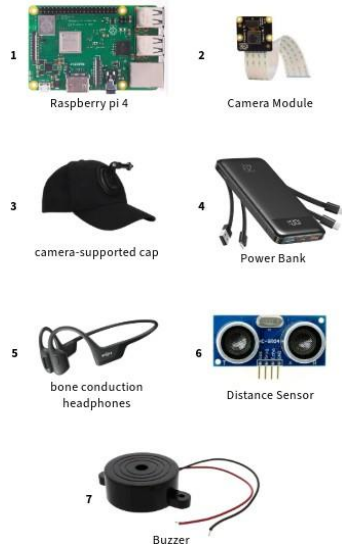

Figure 3: Hardware components of the proposed system system

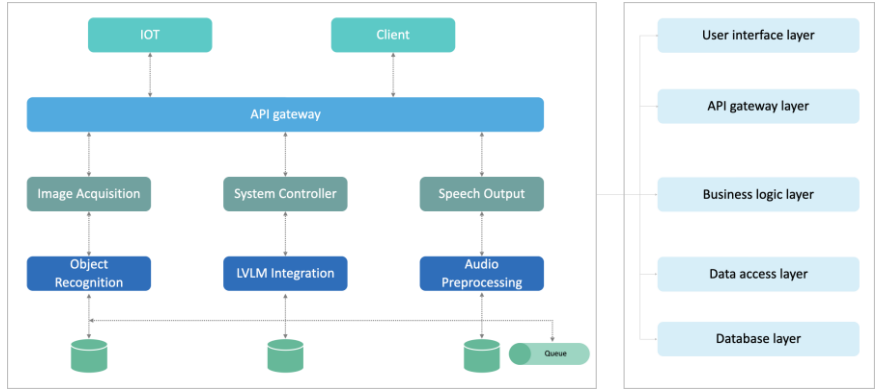

Figure 4: The software architecture of the proposed system

*3.4. Software Architecture*

The system's software architecture includes several modules designed for efficient data processing and user interaction. It is composed of several modules designed for efficient data processing and user interaction. The Raspberry Pi runs on a Linux-based operating system (Raspbian) with real-time processing capabilities. The image acquisition module uses OpenCV [47] to capture images from the camera, manage frame rates, and preprocess images. For object Recognition, the system implements MobileNet SSD [48] for object detection and FaceNet [49] for face recognition, both optimized for the Raspberry Pi's ARM architecture using TensorFlow Lite [50].

To enhance performance, the models are quantized to reduce their size and improve inference speed without sacrificing significant accuracy [51]. A lightweight database, SQLite, is employed for managing the feature embeddings and associated labels to enable quick retrieval during recognition tasks. The system also integrates a large language model through API calls to OpenAI's GPT-4o-mini via Azure. The audio processing module utilizes local text-to-speech systems, such as Festival [52], to convert written text into speech. User inputs are managed through the user interface module, which provides feedback and guides the user through interactions.

*3.5. Integration and Optimization*

For the best performance, the system includes different optimization procedures in its design. As with stitching, critical paradigms are performed in parallel on different threads, allowing efficient usage of the Raspberry Pi quad core CPU and avoiding processing delays. In addition, parallelism uses the GPU to assign some tasks to the Raspberry Pi GPU to increase process- ing speed where necessary. Another feature of the system is resource usage monitoring, which monitors the CPU and memory spare capacity in real time and enables their fine-tuning in real time, thus avoiding system overloading and ensuring stable system work. In addition, to enhance power management, low-power modes of inactive components are incorporated, and dim display mode is enabled during the debugging.

*3.6. Implementation of Object Recognition*

For the object recognition module, captured images were resized to 300x300 pixels and adjusted to meet the input requirements of the MobileNet SSD model. The resized images were then fed into the model to detect and classify objects. During classification Non-Max Suppression technique was used to remove duplicate detections of the same object as well as confidence scores was settled as great than 50% for the reliable detection of objects.

*3.7. Implementation of Face Recognition*

The Haar cascade classifier [53] was used to identify faces and then 128-dimensional embeddings were generated with FaceNet. Finally, the personalized database is compared against embeddings using cosine similarity.

*3.8. Integration with Real-Time Object addition (Green button workflow)*

The system facilitates user interaction by allowing users to capture and label objects with the press of a button, using voice commands or text. The labeled information is then stored for data collection and used to enhance the training dataset. To further enrich this dataset, Generative Adversarial Networks (GANs) are employed to generate synthetic data, which helps supplement the training process. The model is fine-tuned in real-time by incorporating the newly labeled data, ensuring that it adapts to the most recent inputs. Finally, the revised labels are integrated into the model, allowing real-time detection and improving the overall performance of the system.

*3.9. Integration with Large Language Models*

When the 'Blue Request' button is pressed, the system initiates a sequence of actions. It first gathers contextual information by identifying object labels, spatial relationships, and environmental cues. Then, an API call is made, where a prompt is generated and sent to the GPT-4o-mini model using the Ope-nAI API through Azure. The response handling step involves processing the output from the model and formatting it to be suitable for speech conversion. Finally, the system responds to user by delivering the spoken response via bone-conduction headphones, allowing the user to hear the audio while remaining aware of their surroundings. There is strong security system, which guarantee privacy of users and also provide the protection of the information provided by the users. The data minimization principles are adhered to the latter so that only required information is transmitted; in the case of buttons being used by the user to start a process, images are not transmitted to OpenAI unless ignited by a button. All API calls to OpenAI are made securely through Microsoft Azure, the call itself is encrypted to increase the level of security for users.

*3.10. User Interaction Workflow*

The system's interaction with user involves several key steps. It begins with detection and notification, where the system continuously scans the environment and alerts the user to recognized objects and people through brief audio messages. User can then add items by pressing the 'Add Item' button and following the provided voice prompts to include new objects or faces in the system's database. When user presses the 'Detail Request' button, the system responds by delivering a detailed description of the detected object or person.

**4. System Evaluation**

*4.1. Evaluation based on Performance Metrics*

The performance of the employed models has been evaluated using several important evaluation parameters that include Precision, Recall, F-Score, and Accuracy values. Each of these parameter is briefly explained below.

- **Precision**:

  Precision refers to the proportion of the predicted positives of the model that are positive. Mathematically,

  $$Precision = \frac{TP}{TP + FP} \qquad (1)$$

- **Recall**:

  Recall refers to the proportion of the actual positives correctly classified as positives by the model. Mathematically,

  $$Recall = \frac{TP}{TP + FN} \qquad (2)$$

- **F-Score**:

  F-Score is the harmonic mean of the Precision and Recall. Mathematically,

  $$F-Score = \frac{2*TP}{2*TP + FP + FN} \qquad (3)$$

- **Accuracy**:

  Accuracy refers to the proportion of all the correct classifications of the model, Mathematically,

  $$Accuracy = \frac{TP + TN}{TP + TN + FP + FN} \qquad (4)$$

In the above all equations, TP, TN, FP, and FN are the values taken from the confusion matrix. Where;
**TP (True Positives)**: represents the actual positives predicted as positive by the model,
**TN (True Negatives)**: represents the actual negatives predicted as negative by the model,
**FP (False Positives)**: represents the predicted positives by the model, which are negatives,
**FN (False Negatives)**: represents the predicted negatives by the model, which are positives.

*4.2. Evaluation based on Usability Study*

For the usability evaluation of the proposed system, 50 visually impaired participants participants having 25 men and 25 women, aged between 18 and 70 years. Table 1 provides more information about the participants of the usability study, their vision impairment levels and their age groups. As seen from Table 1 that participants with different visual loss levels have been selected to make the study diverse and provide a wide range of experience. The participants tested the device in conditions that were created to resemble real-life situations. We created situations like moving inside, identifying objects in a kitchen, and having conversations in a simulated café and even picking up fruits those they want to eat. This arrangement enabled us to determine how well the device would function in real-life scenarios.

Table 1: Demographic Breakdown of Participants

| **Category** | **Subcategory** | **Count** | **Percentage (%)** |
|---|---|---|---|
| Gender | Male | 25 | 50% |
| | Female | 25 | 50% |
| Age Group | 18–30 | 15 | 30% |
| | 31–50 | 20 | 40% |
| | 51–70 | 15 | 30% |
| Visual Impairment Level | Low Vision | 30 | 60% |
| | Blind | 20 | 40% |

**5. Results Achieved**

*5.1. Recognition Accuracy, Precision, Recall, and F-Score Values*

Results achieved by the proposed system have been listed in Table 2 and also shown through the Confusion Matrix presented in Figure 5. According to Figure 5, when the lighting was good, the object recognition system and LVLM contextual system hit a solid accuracy rate of 90%. This means that 90% of the objects detected by the proposed system were identified correctly. However, there were still some mistakes—10% were false positives, where the system thought it saw an object that wasn't there. In the good lighting situation, the system also achieved desirable values of Precision, Recall, and F-measures (refer to Table 2), indicating the good performance of the system.

In lower light, the accuracy dipped to 80%. In this case, both precision and recall were at 80%, indicating that 80% of the objects detected by the system were correctly identified, while 20% were falsely recognized.

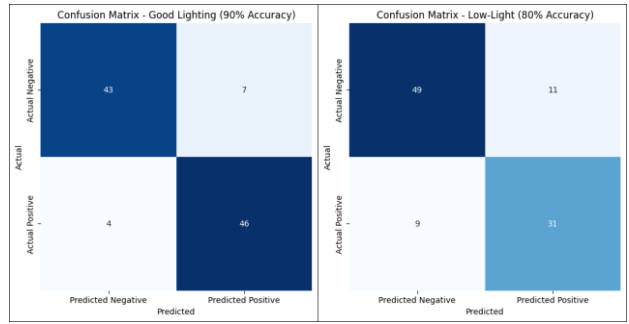

Figure 5: Confusion Matrix of system under Good Lightning and Low Lightning condition

## 5.2. Dark blue button results

When the dark blue button is pressed, the system captures an image, which is then processed by a LVLM to generate detailed information about the image and convey it to the user by voice.

Table 2: Precision, Recall, F1-Score, and Accuracy for Good Lighting and Low-Light Conditions

| Condition | Precision | Recall | F1-Score | Accuracy |
|---|---|---|---|---|
| Good Lighting | 0.88 | 0.92 | 0.90 | 0.90 |
| Low-Light | 0.76 | 0.74 | 0.75 | 0.80 |

assistance. Figures 6 and 7 indicate the system's dark blue button pressing results. It indicate that the system provides the detailed descriptions of the objects by capturing their images, which are then processed by the LVLM to generate their descriptions. The description of the captured images are then communicated to the user using a voice assistance. This way user can obtain a detail description of its surrounding and take better decisions.

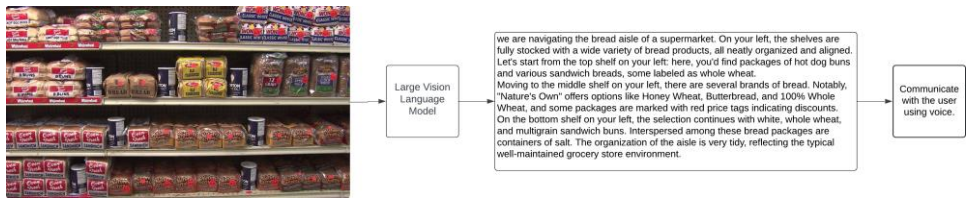

Figure 6: Demo output of Dark blue button (user is standing in bread market)

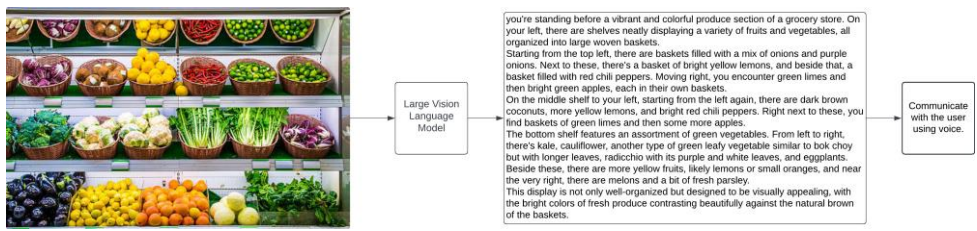

Figure 7: Demo output of Dark blue button (user is standing in vegetable store)

*5.3. User Satisfaction and Usability*

The system demonstrated exceptional user acceptance metrics during evaluation. A shown in Table 3, the System Usability Scale (SUS) assessment yielded an impressive average score of 85 out of 100, indicating high usability. This quantitative success was further reinforced by qualitative user feedback about different features of the systems. Table 4 indicates that over 90% of participants expressed that they found the device helpful and indicated their intention to incorporate it into their regular routines.

*5.4. Response Time*

The system demonstrates efficient performance metrics across different operational modes. In real-time recognition scenarios, the system achieves a swift average response time of 1.5 seconds, enabling immediate object identification. For Detail Requests, which require more comprehensive processing through API calls, the system maintains an average response time of 5.5 seconds, with this additional duration primarily attributed to API call latency.

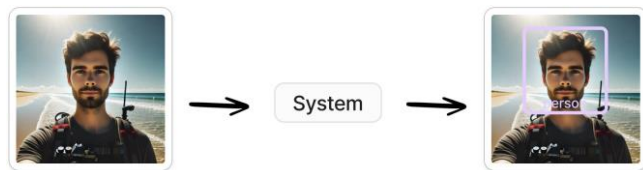

Figure 8: **Core system detecting image as shown in methodology 2**

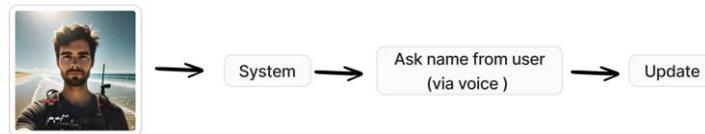

Figure 9: **Label updating process using green button as shown in methodology 2**

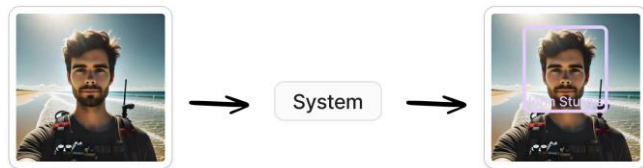

Figure 10: **Using core system again after updating label**

Table 3: Users' Satisfaction Scores

| Scenario | Satisfaction Score (Avg) | Ease of Use (%) | Suggestions |
|---|---|---|---|
| Navigation | 4.5/5 | 90 | Add more precise distance alerts |
| Object Identification | 4.2/5 | 85 | Improve recognition speed |
| Shopping Assistance | 4.0/5 | 80 | Better product categorization |

Table 4: Feature Usability

| Feature | Usage Frequency (%) | Found Very Helpful (%) | Found Less Helpful (%) | User Suggestions |
|---|---|---|---|---|
| Real-Time Object Detection | 9 | 92 | 8 | Add more object categories by default |
| Distance Alert | 90 | 85 | 15 | Increase warning range |
| Personalized Database | 80 | 75 | 25 | Simplify data entry process |

*5.5. Comparative Analysis*

When comparing our system with other conventional devices, the following impressions were made based on the performance analysis of various aspects. Participants consistently reported an enhanced sense of environmental awareness while using the device, indicating superior spatial and contextual understanding. The system's capability to recognize acquaintances proved particularly valuable, leading to improved social comfort and more natural interactions. Furthermore, users expressed a notably increased sense of independence compared to their experience with traditional aids, suggesting that our device successfully addresses key limitations of existing assistive technologies. Table 5 provides the comparative summary of the proposed system with the existing state of the art systems.

Table 5: Comparative summary of the Proposed System with the State of the Art

| Features | Proposed System | AIris [54] | MagicEye [55] | OrCam MyEye [35] | DRISHTI [56] | NewVision [57] |
|---|---|---|---|---|---|---|
| **Object Detection** | Real-time object recognition with LVLMs for enhanced accuracy and contextual insights | Basic object detection for navigation and task assistance | CNN-based model for general object recognition | Recognizes objects via barcodes; less flexible for real-time detection | Detects objects and obstacles for navigation | Utilizes deep learning for object identification and obstacle avoidance |
| **Contextual Understanding** | Deep contextual understanding using large vision-language models (LVLMs) | Scene description provided but lacks in-depth contextual analysis | Limited contextual understanding of object interactions | Provides audio descriptions without comprehensive contextual understanding | Minimal contextual insights; primarily focused on navigation | Integrates voice recognition and assistants for real-time environmental information |
| **Face Recognition** | Personalized face recognition with a user-updated database along with real-time database updation | Limited facial recognition for basic identification tasks | Includes facial recognition | Strong facial recognition capability | Does not support face recognition | Capable of identifying people through facial recognition |
| **Navigation Assistance** | Integrated obstacle detection and personalized audio feedback | Basic navigation support with distance alerts | Provides proximity alerts but lacks advanced navigation feedback | Not designed for navigation | Core functionality with obstacle detection and auditory alerts | Assists in navigation using a combination of sensors and voice commands |
| **Personalization** | Fully user-centric with customizable object and face databases | General-purpose; no personalization options | No database for personalization | No user-updated database; relies on pre-trained recognition data | Limited personalization; navigation is generic | Allows interaction through voice commands for personalized assistance |

# 6. Discussion

## 6.1. Achievements

The proposed system represents a significant advancement in assistive technology by effectively merging cost-effective hardware with cutting-edge AI capabilities. Our implementation achieves accuracy levels comparable to more expensive proprietary systems while maintaining real-time performance that meets the practical demands of everyday use. The system's architecture emphasizes user-centric design through straightforward personalization features, allowing users to easily add new items and expand the system's recognition capabilities. Furthermore, the integration of LLMs enhances the system's contextual understanding, providing users with rich environmental descriptions that enable more independent interactions, including autonomous shopping experiences. This combination of high accuracy, responsive performance, intuitive personalization, and advanced contextual awareness delivers a comprehensive solution that makes sophisticated assistive technology more accessible and practical for daily use.

## 6.2. Limitations

The current implementation faces several technical constraints that warrant consideration. While the Raspberry Pi proves capable for basic operations, it imposes limitations on the complexity of models that can be executed locally, potentially restricting more advanced processing capabilities. Environmental factors also impact system performance, particularly in low-light conditions, suggesting that future iterations could benefit from the integration of infrared capabilities. Additionally, the system's reliance on network connectivity for certain features presents a notable dependency that could affect functionality in areas with limited internet access.

## 6.3. Future Improvements

To further improve system performance and capabilities, several avenues can be explored for future research and development. One promising direction is hardware upgrades, particularly by leveraging more powerful and portable solutions like the NVIDIA Jetson series. This could significantly enhance computational efficiency, making the system more suitable for real-time applications in various environments. Another area of focus should be model optimization. Investigating more efficient algorithms and incorporating custom hardware accelerators can help streamline the processing power required, potentially reducing latency and enhancing overall performance. Additionally, integrating advanced sensors like depth cameras or thermal imaging could improve the system's environmental understanding, particularly in challenging lighting conditions or diverse settings. Enhancing the user interface is also crucial for broader adoption. Introducing features such as voice-activated controls and ensuring compatibility with smartphones would make the sys- tem more user-friendly and accessible to a wider range of users. Lastly, implementing real-time integration with large lan- guage models (LLMs) using high-performance GPUs could fur- ther upgrade the existing setup. By utilizing open-source models, the system could transition from its current configuration, which relies on a dark blue button interface, to one capable of operating seamlessly in real time. This would allow for a more responsive and dynamic interaction experience.

## 7. Conclusion

In this article, we developed a portable vision assistance system that will enhance the mobility and the quality of life of visually impaired people and surprisingly blind and visual impaired people are now able to do shoping and even select things for themself to eat. The idea was to offer practical assistance by combining innovative solutions with simplicity. This system consists of a camera mounted on a hat connected to a Raspberry Pi with an AI for object recognition. It also provides contextual understanding for the first time in research through large language models (LLMs). The system offers real-time audio feedback that helps users to be informed about the environment they are in. This approach is effective in linking conventional assistive devices with modern AI technology, making it affordable and personalized. The practicality of our user testing has shown its effectiveness and the impact it has on users' lives.

Future efforts will focus on addressing current limitations and enhancing the system's capabilities.

## Acknowledgments

The authors extend sincere gratitude to all participants for their valuable time and feedback.

## Funding

This research is not funded by any private/public organization.

## Conflict of Interest Statement

The authors declare that there is no conflict of interest for the publication of this article.

## Data Availability

Data supporting the findings of this study are available from the corresponding author on a reasonable request, with considerations for privacy and ethical compliance.